\documentclass[11pt]{article}

\usepackage[preprint]{acl}

\usepackage{times}
\usepackage{latexsym}

\usepackage[T1]{fontenc}
\usepackage[utf8]{inputenc}

\usepackage{microtype}

\usepackage{inconsolata}

\usepackage{graphicx}
\usepackage{subcaption}
\usepackage{amssymb}
\usepackage{amsmath}
\usepackage{booktabs}
\usepackage{adjustbox}
\usepackage{multirow}
\usepackage{colortbl}
\usepackage[table]{xcolor}

\title{pQuant: Towards Effective Low-Bit Language Models via Decoupled Linear Quantization-Aware Training}

\author{
 \textbf{Wenzheng Zhang\textsuperscript{1}},
 \textbf{Bingzheng Liu\textsuperscript{2}},
 \textbf{Yang Hu\textsuperscript{3}},
 \textbf{Xiaoying Bai\textsuperscript{4}},
 \textbf{Wentao Zhang\textsuperscript{5}},
 \textbf{Bin Cui\textsuperscript{1}},
\\
\\
 \textsuperscript{1}School of Computer Science, Peking University
\\
 \textsuperscript{2}College of Future Information Technology, Fudan University
\\
 \textsuperscript{3}Center for Information Research, Academy of Military Sciences
\\
 \textsuperscript{4}Advanced Institute of Big Data
\\
 \textsuperscript{5}Center of Machine Learning Research, Peking University
\\
}

\begin{document}
\maketitle
\begin{abstract}
  Quantization-Aware Training from scratch has emerged as a promising approach for building efficient large language models (LLMs) with extremely low-bit weights (sub 2-bit), which can offer substantial advantages for edge deployment. However, existing methods still fail to achieve satisfactory accuracy and scalability. In this work, we identify a parameter democratization effect as a key bottleneck: the sensitivity of all parameters becomes homogenized, severely limiting expressivity. To address this, we propose pQuant, a method that decouples parameters by splitting linear layers into two specialized branches: a dominant 1-bit branch for efficient computation and a compact high-precision branch dedicated to preserving the most sensitive parameters. Through tailored feature scaling, we explicitly guide the model to allocate sensitive parameters to the high-precision branch. Furthermore, we extend this branch into multiple, sparsely-activated experts, enabling efficient capacity scaling. Extensive experiments indicate our pQuant achieves state-of-the-art performance in extremely low-bit quantization.
\end{abstract}

\section{Introduction}
\label{Introduction}
Large Language Models (LLMs) ~\cite{touvron2023llama, yang2024qwen2, team2023gemini, achiam2023gpt} have achieved remarkable performance across diverse natural language processing tasks. However, their massive computational and memory demands pose significant challenges for practical deployment, especially on resource-constrained devices ~\cite{dettmers2022gpt3}.  Quantization, which reduces the numerical precision of model weights and activations, has become a key technique to mitigate these constraints ~\cite{frantar2022gptq, yao2021hawq, liu2023llm, yuan2023rptq, zhang2025pqcache}. 

Extremely low-bit (sub 2-bit) quantization stands out for offering the most aggressive model compression, promising drastic reductions in memory and compute footprint ~\cite{qin2022bibert, shang2023pb, huang2024billm, ma2024fbi, xu2024onebit}. Its key hardware advantage lies in replacing costly floating-point matrix multiplications with efficient bitwise operations during inference, presenting a promising path toward edge-deployable LLMs ~\cite{zhu2024scalable}. Recent advances in extremely low-bit quantization have yielded notable empirical results. For instance, post-training quantization (PTQ) methods such as PTQ1.61 ~\cite{zhao2025ptq1} and BiLLM ~\cite{huang2024billm} achieve 68.5\% and 61.2\%, respectively, of FP16 baseline accuracy on downstream tasks. Quantization-aware fine-tuning further improves performance, with state-of-the-art approaches recovering up to 79.3\% of FP16 accuracy ~\cite{xu2024onebit}. 

\begin{figure}[t]
  \begin{center}
  \centerline{\includegraphics[width=0.85\columnwidth]{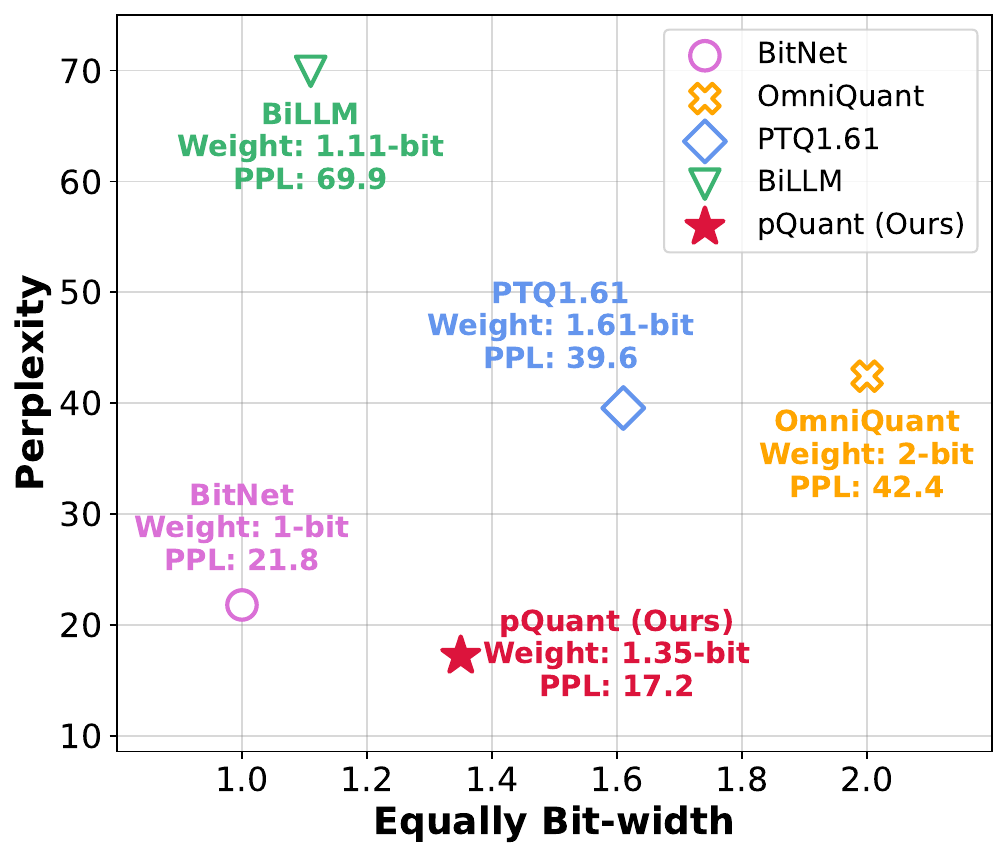}}
  \caption{An overview of performance (Perplexity on WikiText2) and bit-width achieved by our pQuant and other extremely low-bit methods on 1.3B model.}
  \label{fig:intro_scaling_law}
  \end{center}
\end{figure}

Despite their strong efficiency benefits, extremely low-bit LLMs face limited practical adoption due to the accuracy degradation, less than 80\% of FP16 capability. This challenge has spurred interest in Quantization-Aware Training from scratch (QAT-Scratch), which trains low-bit models directly from random initialization without compressing pre-trained weights ~\cite{wang2023bitnet, ma2024fbi, ma2025bitnet}. QAT-Scratch shows strong promise: BitNet achieves 90.1\% FP16 parity on downstream tasks ~\cite{wang2023bitnet}, and BitNet1.58 enables near-lossless 2-bit quantization with a 3B model ~\cite{ma2024era}. Nevertheless, our experiments reveal two critical issues that persist in current 1-bit QAT-Scratch models: (i) even the state-of-the-art 1-bit method still exhibit a non-negligible accuracy gap; and (ii) scaling efficiency is poor: as model size increases, performance gains grow sublinearly and fall far behind those of FP16 models. This limits the practicality of extremely low-bit LMs in large-scale applications.

What underlies this performance and scaling bottleneck? It is well-established in quantization theory that parameters possess varying sensitivity, with a small subset of "sensitive" parameters disproportionately influencing model output ~\cite{dettmers2022gpt3, dettmers2023spqr, lee2024owq, li2025lut}. Surprisingly, we find that existing 1-bit QAT-Scratch models lose this structure: weight sensitivity becomes nearly uniform. We term this phenomenon parameter democratization, defined as the unintended homogenization of parameter sensitivity under extreme quantization. This observation motivates our central hypothesis: parameter democratization may be a key factor limiting the expressivity and scalability of 1-bit LMs.

To address and empirically validate our hypothesis on parameter democratization, we propose pQuant, a novel extremely low-bit QAT-Scratch method founded on the principle of decoupled and guided parameter sensitivity. The core of pQuant is a decoupled linear layer that structurally enforces parameter specialization: a 1-bit main branch ensures foundational efficiency, while a compact high-precision branch is dedicated to preserving sensitive parameters. Critically, importance assignment is not predetermined. Instead, feature scaling guide the model to dynamically allocate its most influential representations to the high-precision pathway. Build upon this, we further evolve the high-precision branch into a sparsely activated expert module, where a lightweight router selects one expert per token. This design enables substantial capacity growth with near-constant inference cost. Experiments show that pQuant reduces perplexity by 32.0\% over state-of-the-art 1-bit baselines, and the high-precision branch effectively preserves the most sensitive parameters. When scaled, it surpasses 2-bit models in accuracy while improving inference throughput by 18.2\%. Notably, it matches the performance of FP16 models while delivering over 2× higher throughput.

\section{Preliminary}
\label{Preliminary}

\subsection{Quantization Aware Training}
Quantization-Aware Training improves model robustness to quantization by simulating low-bit operations during training. Standard QAT fine-tunes a pre-trained full-precision model, using FP16 shadow weights to compute gradients while quantizing weights and activations in the forward pass ~\cite{shen2021once}. Recent advances enhance this paradigm: LLM-QAT ~\cite{liu2023llm} employs data-free distillation; BitDistiller ~\cite{du2024bitdistiller} introduces asymmetric clipping; EfficientQAT ~\cite{chen2024efficientqat} reduces cost via staged optimization; GETA ~\cite{qu2025automatic} jointly optimizes pruning and quantization. Technical details of QAT are provided in Appendix~\ref{appendix:Quantization-Aware Training}. While these methods achieve strong results at 4–8 bit, they face fundamental challenges when pushed to sub-2-bit regimes, often struggling to recover accuracy due to the difficulty of compressing pre-trained high-precision representations.

\subsection{Extremely Low-Bit Quantization}
Extremely low-bit (sub 2-bit) quantization aims for the most hardware-efficient LLMs but suffers from severe accuracy degradation. Early binarization methods like BNN ~\cite{hubara2016binarized} and XNOR-Net ~\cite{rastegari2016xnor} (which introduced scaling factors) do not scale to modern LLMs. Recent work has adapted quantization for LLMs: PB-LLM ~\cite{shang2023pb} identifies salient weights, BiLLM ~\cite{huang2024billm} proposes a two-stage binarization framework, OneBit ~\cite{xu2024onebit} uses SVID decomposition, and BinaryMoS ~\cite{jo2024mixture} employs token-adaptive scales. PTQ1.61 ~\cite{zhao2025ptq1} introduces a one-dimensional structured mask to reduce the upper bound of quantization errors,

A promising direction is Quantization-Aware Training from scratch (QAT-Scratch), which trains low-bit models from random initialization. BitNet ~\cite{wang2023bitnet} pioneered this paradigm for 1-bit LLMs, showing significantly better performance than PTQ or fine-tuning-based QAT. BitNet1.58 ~\cite{ma2025bitnet} extended this success to 2-bit, achieving near-lossless accuracy. Other works like FBI-LLM ~\cite{ma2024fbi} (using progressive distillation) and iFairy ~\cite{wang2025ifairy} (using complex-valued representations) further explore this training paradigm. However, as identified in our work, existing QAT-Scratch methods still face a fundamental bottleneck in expressivity and scalability, which we attribute to the parameter democratization effect.

\subsection{Sensitivity Analysis}
Not all parameters in a neural network contribute equally to its output. Intuitively, a weight is considered sensitive to quantization if it incurs a large rounding error (i.e., it lies far from a quantization grid point) and/or if it frequently multiplies large input activations, thereby amplifying even small errors ~\cite{dettmers2023spqr, lee2024owq, shang2023pb, li2025mbq}.

Inspired by SPQR~\cite{dettmers2023spqr}, we adopt a perturbation-based metric to analyze such sensitivity under extreme low-bit settings. For a weight $w_{ij}$ in matrix $W$ and given calibration inputs $X$, we define its sensitivity as the minimum increase in squared output distortion:
\begin{equation}
  s_{ij} = \min_{W'} \|WX - W'X\|_2^2,
\end{equation}
where $W'$ satisfies $w'_{ij} = \mathrm{quant}(w_{ij})$, and all other entries of $W'$ are free to adjust for optimal error compensation. Crucially, this problem admits a closed-form solution under the generalized Optimal Brain Surgeon framework~\cite{frantar2022optimal}:
\begin{equation}
    s_{ij} = \frac{w_{ij}^2}{2(XX^\top)^{-1}},
    \label{eq:sensitivity}
\end{equation}
where $(XX^\top)^{-1}$ is the inverse Hessian matrix corresponding to the optimization problem. This term quantifies how "replaceable" the weight is given the correlation structure of the inputs. Thus, sensitivity $s_{ij}$ captures a joint effect: a large rounding error $w_{ij}^2$ may still yield low sensitivity if the weight lies in a direction where the input allows easy compensation. This metric can be efficiently approximated by quantization solvers, such as GPTQ\cite{frantar2022gptq}.

\begin{figure}[t]
  \begin{center}
  \centerline{\includegraphics[width=\columnwidth]{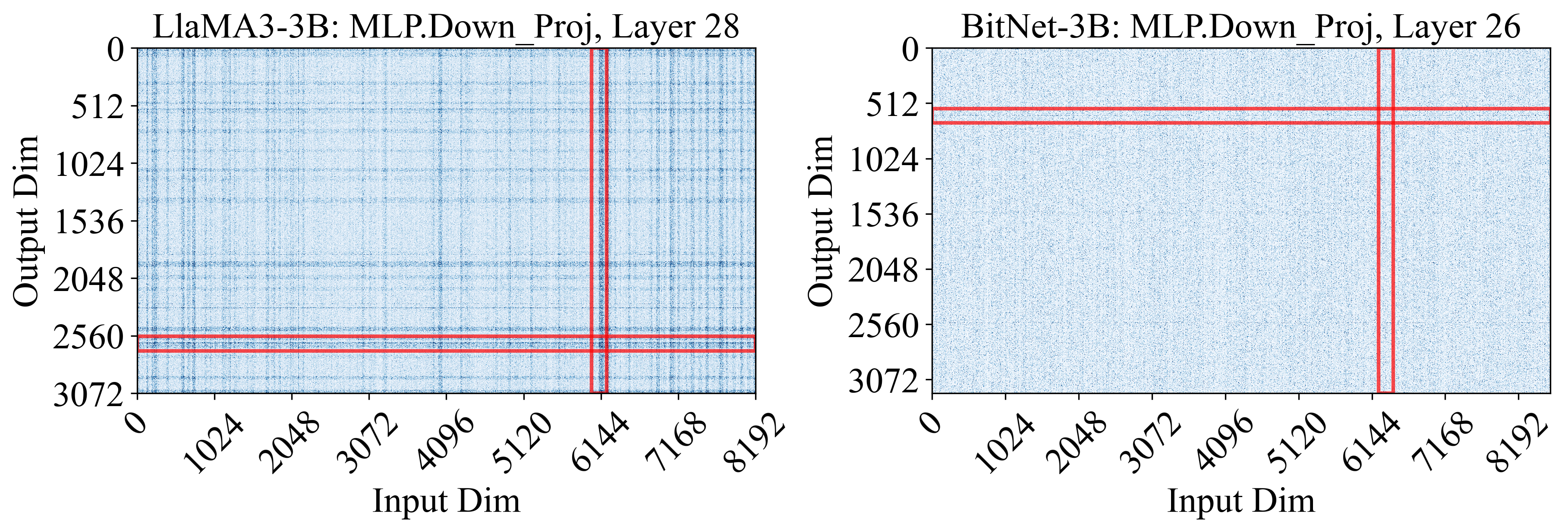}}
  \caption{Weight log-sensitivities in the final FFN layer of LLaMA3-3B and BitNet-3B. Matrices are down sampled via max pooling for visualization; darker blue indicates higher sensitivity. Red boxes highlight regions of peak sensitivity. Notably, in the 1-bit weights of BitNet-3B, no pronounced sensitivity variation is observed.}
  \label{fig:sensitivity_analysis}
  \end{center}
\end{figure}

To analyze the sensitivity landscape under extreme quantization, we set $\mathrm{quant}(w_{ij}) = 0$ to perturb and compute per-layer sensitivity using a small calibration set $X$. As shown in \autoref{fig:sensitivity_analysis}, this reveals a stark contrast: in the 16-bit LLaMA-3 model, a small subset of parameters exhibits significantly higher sensitivity, indicating their disproportionately large influence on the output. In contrast, the 1-bit BitNet model shows a flattened, near-uniform distribution, suggesting that all parameters are treated with comparable importance.

We refer to this observed collapse of sensitivity differentiation as parameter democratization. This homogenization may prevent the model from effectively prioritizing the most informative features. We therefore hypothesize that recovering a differentiated sensitivity structure, one in which critically important parameters are distinctly identified and preserved, could be key to improving both the accuracy and scalability of extremely low-bit LLMs.

\begin{figure*}[th]
    \centering
    \subcaptionbox{The architecture of the decoupled design in pQuant}{\includegraphics[width = 0.98\linewidth]{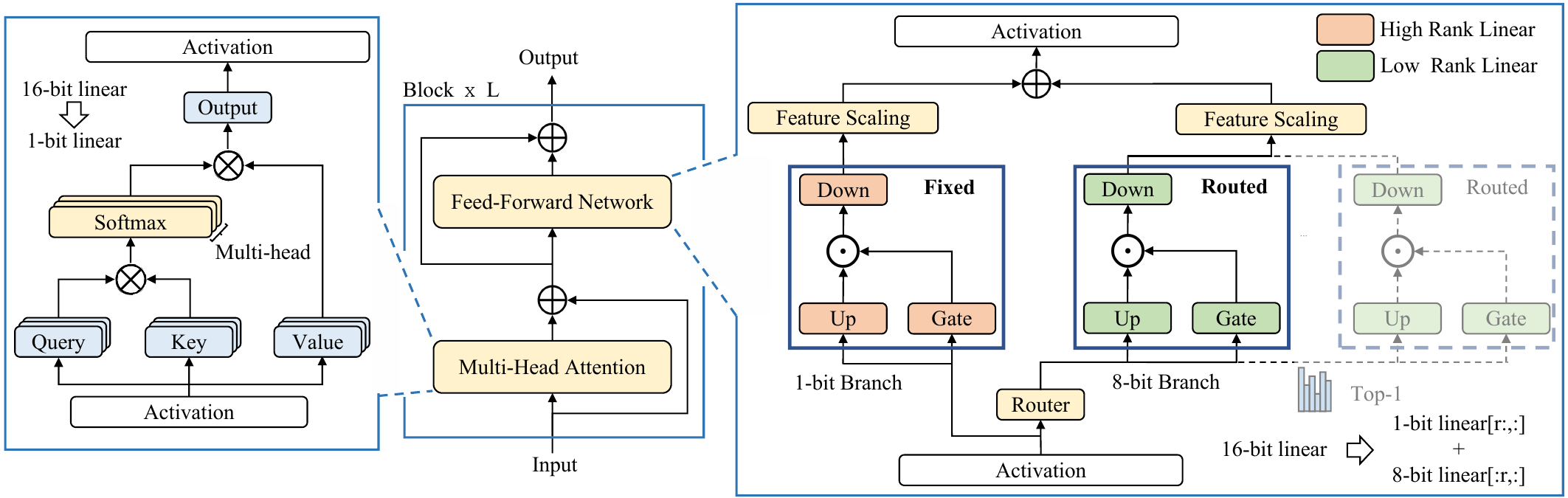}} \\

    \subcaptionbox{1-bit Liner in MHA}{\includegraphics[width = 0.58\columnwidth]{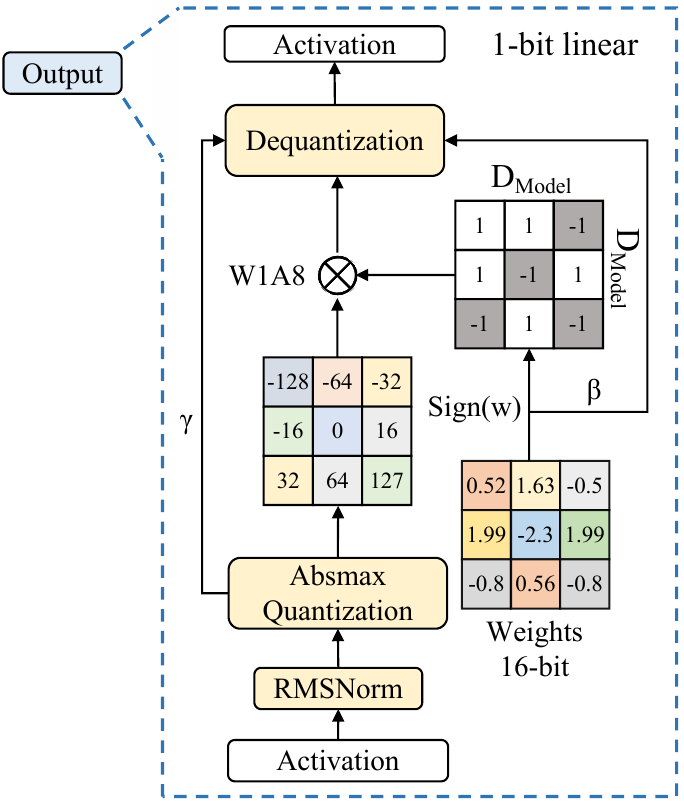}}
    \subcaptionbox{8-bit Linear in FFN}{\includegraphics[width = 0.67\columnwidth]{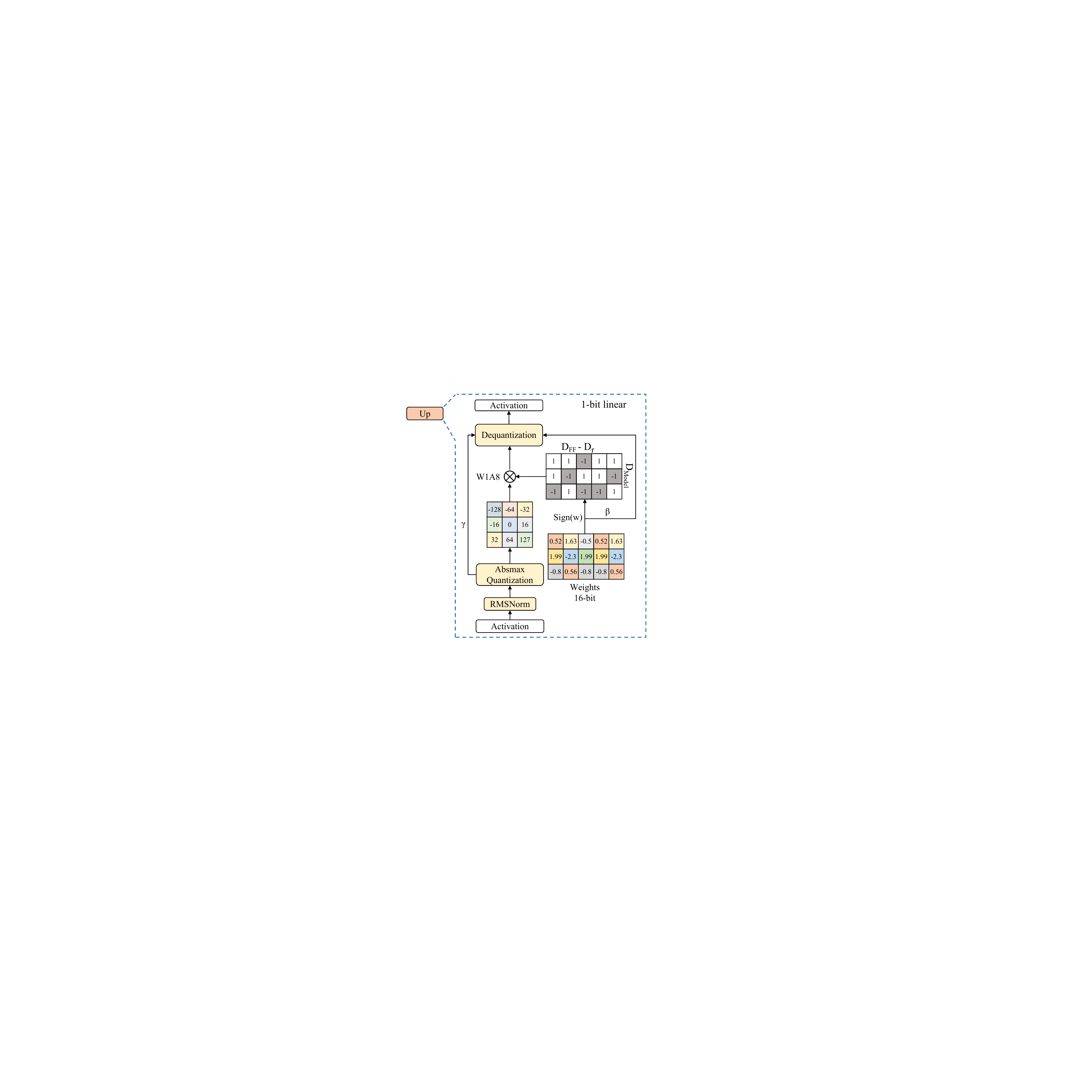}}
    \subcaptionbox{1-bit Linear in FFN}{\includegraphics[width = 0.554\columnwidth]{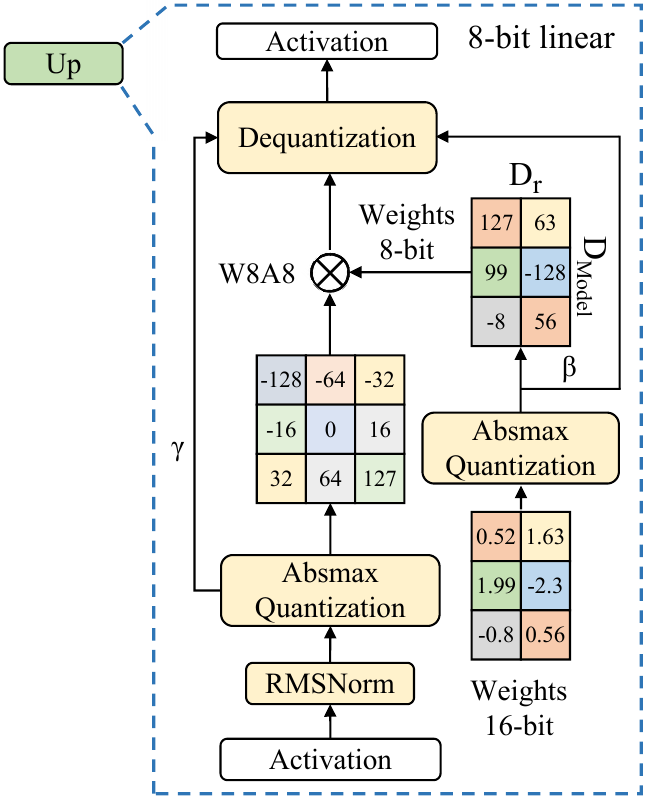}}
    \caption{Computational flow of pQuant’s core modules. (a) pQuant replaces all linear layers with quantized counterparts, with 8-bit branch to enable dynamic scaling when needed. (b-d) Computation in three representative pQuant modules. FP16 weights are retained solely during training to ensure numerical stability and discarded post-training. Here, $r$ denotes the dimension of weights in 8-bit branch, where $r \ll D_{\text{model}}$.}
    \label{fig:architecture}
\end{figure*}

\section{Methodology}
\label{Methodology}
Based on the preceding analysis, we propose pQuant, a method designed to counteract parameter democratization by structurally decoupling parameters. pQuant isolates a small set of sensitive weights into a higher-precision branch within an efficient 1-bit backbone. As illustrated in  \autoref{fig:architecture}, pQuant replaces standard linear layers in both multi-head attention (MHA) and feed-forward network (FFN) modules with quantization-native layers and uses feature scaling to explicitly guide importance-aware learning. 

Within the MHA module, pQuant substitutes the query, key, value, and output projections with a pure 1-bit linear layer (\autoref{fig:architecture}(b)). For the FFN, we design a decoupled linear layer: most parameters are quantized to 1-bit for efficiency (\autoref{fig:architecture}(c)), while a small subset is kept in 8-bit to preserve highly sensitive weights (\autoref{fig:architecture}(d)).

This decoupled design can be naturally viewed through the lens of a Mixture of Experts (MoE): the 1-bit branch acts as a shared expert common to all inputs, ensuring efficiency, while the 8-bit branch functions as a specialized, routable expert. A lightweight, fixed top-1 router dynamically activates routable expert based on the input token.

\subsection{Multi-Head Attention in pQuant}
\label{1-bit Attention}
In the MHA, each linear layer utilizes 1-bit weights represented with INT1. To maximize inference efficiency, we apply this aggressive, undifferentiated quantization specifically to MHA, while reserving our decoupled architecture for the FFN. This targeted design is supported by two key observations from prior work. First, FFN layers are known to concentrate a larger fraction of sensitive parameters that are critical to model performance ~\cite{dettmers2023spqr}. Second, activation distributions in FFNs tend to be less regular and exhibit more outliers compared to the relatively structured activations in attention layers ~\cite{shazeer2020glu}.

During pre-training, weights and activations are dynamically quantized to low precision, while gradients and optimizer states are maintained in FP32 to preserve training stability and accuracy. Specifically, FP16 weights are used in the training pass to ensure meaningful gradient updates. Using 1-bit weights during optimization would result in near-zero gradients for most parameters, making it difficult for the model to learn effective parameter updates and reduce the loss. These weights are discarded during inference, leaving only the 1-bit parameters in the model. 

These FP16 weights are quantized to 1-bit before performing the tensor multiplication between weights and activations. After the multiplication, a dequantization process is applied to the output. This process can be formally described as follows:
\begin{equation}
W^{\mathit{INT1}} = Sign( W^{\mathit{Float}} - \mu )
\end{equation}
\begin{equation}
    Sign(W_{ij}) =
    \begin{cases}
    +1, & \text{if } W_{ij} > 0 \\
    -1, & \text{if } W_{ij} < 0
    \end{cases}
\end{equation}
\begin{equation}
Y = \lambda \cdot W^{\mathit{INT1}} LayerNorm(X)
\label{Output of dequantization} 
\end{equation}
\begin{equation}
  \mu = \tfrac{1}{mn} \!\sum_{i=1}^{m} \!\sum_{j=1}^{n} \!W_{ij},
  \lambda = \tfrac{1}{mn} \!\sum_{i=1}^{m} \!\sum_{j=1}^{n} \!|W_{ij}|
  \label{eq:mu_lambda}
\end{equation}

Here, $W^{\mathit{INT1}} \in \mathbb{R}^{m \times n}$ denotes the 1-bit quantized weight matrix, and $W^{\text{float}}$ the FP16 reference. $X$ and $Y$ are the input and output of the linear layer, respectively. To reduce the $\ell_2$ quantization error, a dequantization scale $\lambda$ is applied. Following \citet{liu2022bit}, we center the weights to zero mean using $\mu$ prior to binarization, which enhances the information capacity of binary weights. Additionally, pQuant employ the AbsMax method~\cite{dettmers2024qlora} to quantization all activations and high-precision branch in FFN along the token dimension to the range $[-2^{7}, 2^{7}]$, and represent them in INT8 format:
\begin{equation}
Q(X) = RoundClip(X \times \gamma, -2^{7}+\epsilon, 2^{7}+\epsilon)
\end{equation}
\begin{multline}
  RoundClip(X, a, b)  \\
  = \max(a,\min(b,Round(X)),
\end{multline}
\begin{equation}
\gamma =  \frac{2^{7}}{\max(|x_1|, |x_2|, \dots, |x_n|)}
\end{equation}
where $Round(\cdot)$ function rounds each value to the nearest integer. The parameter $\epsilon$ is a small floating-point value that prevents overflow during clipping.  Additionally, $\gamma$ is the scaling factory for activations, which will be incorporated with $\lambda$ to de-quantize the output as follows:
\begin{equation}
Y = \frac{\lambda}{\gamma} \times W^{INT1} Q(LayerNorm(X))  
\end{equation}

\subsection{Feed-Forward Network in pQuant}
\label{FFN in pQuant}
To preserve highly sensitive parameters with higher precision, a straightforward approach is to designate a subset of dimensions in the linear layer to be represented in high precision while keeping the rest in 1-bit. However, this naïve strategy faces two inherent limitations in the QAT-Scratch setting. First, since parameters are randomly initialized and trained from scratch, there exists no consistent spatial pattern in the weight matrix that reliably indicates where sensitive parameters reside \cite{yu2024super}. Predefining fixed positions for high-precision weights thus diminishes the potential benefit. Second, the non-uniform distribution of input activation magnitudes can dilute the advantage of retaining higher precision in arbitrary dimensions, often rendering the improvement marginal—similar to random selection. These issues highlight the need for a more adaptive and learnable strategy for high-precision allocation.

Accordingly, we introduce the decoupled linear layer, which structurally splits the original weight matrix in the FFN into two parallel computational branches: a 1-bit branch and a high-precision branch. This design allows the high-precision branch to operate independently, effectively avoiding the dynamic range collapse issue inherent in per-tensor 1-bit quantization and AbsMean scaling. Both branches process the same input activations and their outputs are then summed to form the final result.

In pQuant, we adopt 8-bit precision for the high-precision branch, motivated by extensive evidence that 8-bit representations preserve essential parameter information while maintaining computational efficiency~\cite{liu2024deepseekv3, peng2023fp8, van2023fp8, mishra2025recipes}. Among 8-bit formats, we specifically choose INT8 over FP8 or MXFP8 due to its superior hardware support and deployment compatibility on mainstream CPU/GPU platforms.

A naïve summation of the two branches’ outputs would conflate their distinct sensitivity profiles, hindering the model’s ability to learn their relative importance. Inspired by RSLoRA and MS-LoRA~\cite{kalajdzievski2023rank, luo2025mslora}, which demonstrate that scaling informative components enhances representational capacity allocation. We introduce \emph{feature scaling}, where the outputs of the two FFN branches are modulated by learnable scalars $\alpha$ and $\beta$. To prioritize the high-precision signal during optimization, we initialize $\alpha \gg \beta$ (e.g., $\alpha = 2.0$, $\beta = 0.2$), ensuring the 8-bit branch receives stronger gradient feedback during back propagation. The FFN computation is then:
\begin{multline}
  Y = \alpha\, \mathit{FFN}_{[:r]}^{\mathit{INT8}}(\mathit{LayerNorm}(X)) \\
  + \beta\, \mathit{FFN}_{[r:]}^{\mathit{INT1}}(\mathit{LayerNorm}(X)),
  \label{eq:y_multline}
\end{multline}
where $r$ is a hyperparameter controlling the fraction of high-precision weights retained to preserve sensitivity-critical parameters, with $r \in [0, D_{\text{ffn}}]$ and $D_{\text{ffn}}$ denoting the FFN hidden dimension. The 1-bit weights follow the same quantization scheme as in the MHA’s 1-bit linear layers (\S\ref{1-bit Attention}), while the 8-bit weights are quantized identically to 8-bit activations (\autoref{1-bit Attention}).

\subsection{Efficient Scaling of pQuant}
We expand the high-precision branches into multiple parallel branches and employs a lightweight router to dynamically activate the most suitable path based on the input features. This structure serves as an optional mechanism for performance enhancement. Specifically, pQuant extend the number of 8-bit branch to $N$, incorporating a gating to select the most appropriate weights for input activations, as shown in \autoref{fig:architecture}(a). The resulting FFN architecture bears a resemblance to Mixture-of-Experts (e.g., DeepSeekMoE ~\cite{dai2024deepseekmoe} and OpenMoE ~\cite{xue2024openmoe}). However, the two designs are fundamentally different due to the discrepancy in precision and dimension. 

The router in pQuant is implemented as a simple linear layer and trained end-to-end along with the rest of the model. pQuant adopts a top-1 gating strategy to maximize computational efficiency, ensuring that the number of active parameters in the FFN matches that of a conventional dense layer. The gating function employs a softmax activation~\cite{shazeer2017outrageously}. The $N$ acts as a hyperparameter, with detailed analysis presented in (\S\ref{experiments:Scalability of pQuant}).

\begin{table}[t]
\begin{center}
\begin{adjustbox}{width=\linewidth}
\begin{tabular}{cccccc}
\toprule
\textbf{Parameter} & $\textbf{D}_{\textbf{Model}}$ & $\textbf{D}_{\textbf{FF}}$ & \textbf{r} & $\textbf{1-bit}$ & $\textbf{8-bit}$ \\
\midrule
300M & 1024  & 2272(2400-128) & 128 & 96\% & 4\% \\
700M & 1536  & 3840(4096-256) & 256 & 96\% & 4\% \\
1.3B & 2048  & 5076(5400-384) & 384 & 95\% & 5\% \\
2.6B & 2880  & 7168(7680-512) & 512 & 95\% & 5\% \\
\bottomrule
\end{tabular}
\end{adjustbox}
\end{center}
\caption{Configurations for model trained by pQuant. $\textbf{D}_{\textbf{FF}}$ denote the hidden dimension of the feed-forward network; r denotes the dimension of the 8-bit branch; 1-bit and 8-bit denote the percentage of total parameters represented by 1-bit and 8-bit respectively.}
\label{pQuant configuration}
\end{table}

\section{Experiments}
\label{Experiments}
In this section, we conduct extensive experiments to validate our method pQuant. We further evaluate the inference efficiency. Finally, we show the ablation studies of pQuant.

pQuant is trained end-to-end via QAT-Scratch, adopting a two-stage schedule to ensure stable and robust convergence~\cite{ma2024era}. Gradients for quantized weights during training are approximated using the Straight-Through Estimator (STE)~\cite{bengio2013estimating} during back propagation. Training details are provided in \autoref{appendix:training}. We also optimize the inference efficiency of pQuant, as detailed in \autoref{appendix:Inference}.

\subsection{Experimental setup}
\label{Experimental Setup}
In our experimental, we train pQuant at four scales detailed in \autoref{pQuant configuration}. For a comprehensive comparison, we also train from scratch three representative baselines under identical configurations: BitNet (1-bit model) \cite{wang2023bitnet}, BitNet1.58 (2-bit model) \cite{ma2024era}, and LLaMA-3 (16-bit model) \cite{touvron2023llama2}. Additionally, we compare pQuant against post-training quantization (PTQ) methods, including PTQ1.61 \cite{zhao2025ptq1} and OmniQuant \cite{shao2023omniquant}, as well as fine-tuning-based QAT approaches such as OneBit \cite{xu2024onebit}. Our models are trained on three datasets: C4~\cite{raffel2020exploring}, Wikipedia~\cite{wikidump} and ArXiv~\cite{cohan-etal-2018-discourse}. We configured the batch size to 4 million tokens for LlaMA-2 and 1 million tokens for the other models.

We evaluate the models based on their zero-shot performance in some downstream tasks, including ARC-Easy (ARC-E) ~\cite{yadav2019quick}, ARC-Challenge (ARC-C) ~\cite{yadav2019quick}, BoolQ(BQ)~\cite{clark2019boolq}, PIQA(PQ)~\cite{bisk2020piqa}, Winogrande(WGe)~\cite{sakaguchi2021winogrande}, OpenbookQA(OQ)~\cite{mihaylov2018can}, and Hellaswag(HS)~\cite{zellers2019hellaswag}. More details about experimental setup is shown in \autoref{Experiment Details}.

\begin{table*}[th]
\begin{center}
\begin{adjustbox}{width=\textwidth}
\begin{tabular}{cccccccccccc}
\toprule
\textbf{Parameters} & \textbf{Method} & \textbf{Bit} & \textbf{ARC-E$\uparrow$} & \textbf{ARC-C$\uparrow$} & \textbf{HS$\uparrow$} & \textbf{BQ$\uparrow$} & \textbf{OQ$\uparrow$} & \textbf{PQ$\uparrow$} & \textbf{WGe$\uparrow$} & \textbf{Avg.$\uparrow$} & \textbf{Perplexity}$\downarrow$ \\
\midrule
\multirow{3}{*}{300M} & FP16 & 16 & 44.8 & 19.3 & 34.8 & 60.7 & 18.6 & 66.8 & 51.3 & 42.3 & 22.9 \\ \cmidrule{2-12}
  & BitNet & 1 & 38.2 & 16.2 & 32.9 & 53.5 & 14.2 & 58.1 & 48.8 & 37.6 & 34.2 \\
& \cellcolor{gray!15}\textbf{pQuant} & \cellcolor{gray!15}1.28 & \cellcolor{gray!15}41.5 & \cellcolor{gray!15}19.7 & \cellcolor{gray!15}33.0 & \cellcolor{gray!15}60.0 & \cellcolor{gray!15}16.2 & \cellcolor{gray!15}62.5 & \cellcolor{gray!15}50.9 & \cellcolor{gray!15}40.5 & \cellcolor{gray!15}30.1 \\
\midrule
\multirow{3}{*}{700M} & FP16 & 16 & 51.4 & 21.4 & 37.3 & 58.9 & 19.0 & 68.5 & 52.4 & 44.1 & 16.5 \\ \cmidrule{2-12}
  & BitNet & 1 & 44.4 & 18.7 & 35.4 & 56.5 & 14.9 & 62.9 & 50.0 & 40.4 & 27.6 \\
& \cellcolor{gray!15}\textbf{pQuant} & \cellcolor{gray!15}1.28 & \cellcolor{gray!15}47.0 & \cellcolor{gray!15}20.1 & \cellcolor{gray!15}36.2 & \cellcolor{gray!15}57.2 & \cellcolor{gray!15}17.4 & \cellcolor{gray!15}67.1 & \cellcolor{gray!15}51.6 & \cellcolor{gray!15}42.4 & \cellcolor{gray!15}21.9 \\
\midrule
\multirow{6}{*}{1.3B} & FP16 & 16 & 53.9 & 22.1 & 38.7 & 55.7 & 21.8 & 71.0 & 54.3 & 45.4 & 14.4 \\ \cmidrule{2-12}
  & BitNet & 1 & 47.8 & 19.2 & 36.3 & 58.6 & 17.6 & 67.6 & 52.0 & 42.6 & 21.8 \\
  & BitNet1.58 & 2 & 50.4 & 21.2 & 36.9 & 61.2 & 19.4 & 68.7 & 53.0 & 44.4 & 16.9 \\
  & OmniQuant & 2 & 38.8 & 23.4 & 33.4 & 56.5 & 15.3 & 60.9 & 51.9 & 40.0 & 42.43 \\
  & OneBit & 1.1 & 41.3 & 24.1 & 34.3 & 59.5 & 16.3 & 62.6 & 51.1 & 41.3 & 25.4 \\
& \cellcolor{gray!15}\textbf{pQuant} & \cellcolor{gray!15}1.35 & \cellcolor{gray!15}49.8 & \cellcolor{gray!15}20.8 & \cellcolor{gray!15}\textbf{37.0} & \cellcolor{gray!15}60.2 & \cellcolor{gray!15}\textbf{20.1} & \cellcolor{gray!15}68.3 & \cellcolor{gray!15}52.5 & \cellcolor{gray!15}44.0 & \cellcolor{gray!15}17.2 \\
\midrule
2.6B & \cellcolor{gray!15}\textbf{pQuant} & \cellcolor{gray!15}1.35 & \cellcolor{gray!15}54.6 & \cellcolor{gray!15}22.9 & \cellcolor{gray!15}40.5 & \cellcolor{gray!15}61.6 & \cellcolor{gray!15}22.0 & \cellcolor{gray!15}71.9 & \cellcolor{gray!15}56.4 & \cellcolor{gray!15}47.1 & \cellcolor{gray!15}13.0 \\
\midrule
7B & PTQ1.61 & 1.61 & 47.2 & 22.3 & 35.8 & 56.5 & 15.3 & 63.2 & 52.3 & 41.8 & 12.7 \\
\bottomrule
\end{tabular}
\end{adjustbox}
\caption{Main experimental results. We report perplexity and zero-shot accuracy of BitNet (1-bit), BitNet1.58 (2-bit), FP16 LLaMA-2 (16-bit), and pQuant on downstream tasks. Perplexity is evaluated on WikiText-2. Results show that pQuant significantly improves the performance of 1-bit models.}
\label{tab:eval_results}
\end{center}
\end{table*}

\subsection{Performance of pQuant}
\label{Experimental Results}
The main evaluation results are summarized in \autoref{tab:eval_results}. To strictly isolate architectural contributions from data and scale effects, our primary comparison focuses on BitNet, the most relevant 1-bit QAT-from-scratch baseline, under identical model sizes and data budgets (100B tokens). In this controlled setting, pQuant ($1.28\sim1.35$-bit) achieves consistent gains over BitNet across all metrics. Notably, the 700M-parameter pQuant matches the performance of the larger 1.3B BitNet, suggesting that preserving sensitive weights in a high-precision branch significantly enhances parameter efficiency. 

At the matched 1.3B scale, pQuant narrows the gap to the 2-bit BitNet1.58 to just 0.4 average accuracy points despite using 0.65 fewer bits per weight. We further include comparisons with OmniQuant, OneBit, and PTQ1.61 not as direct competitors under equal conditions, but to highlight the efficacy of QAT-from-scratch against methods leveraging substantially more resources. Specifically, OmniQuant and OneBit are derived from OPT-1.3B pre-trained on 180B tokens (80\% more than our dataset), while PTQ1.61 applies post-training quantization to a 7B LLaMA-2 model trained on $\ge$3T tokens. Despite these disparities, pQuant surpasses both OmniQuant and OneBit at the same scale. Most strikingly, the 2.6B pQuant achieves an average accuracy of 47.1, outperforming the 1.3B LLaMA-2 (FP16) and substantially exceeding PTQ1.61 applied to the much larger 7B model. These results underscore that well-designed low-bit quantization trained from scratch can achieve superior performance even with significantly fewer parameters and pretraining data.

\subsection{Scalability of pQuant}
\label{experiments:Scalability of pQuant}
\begin{figure}[t]
  \begin{center}
  \centerline{\includegraphics[width=0.9\columnwidth]{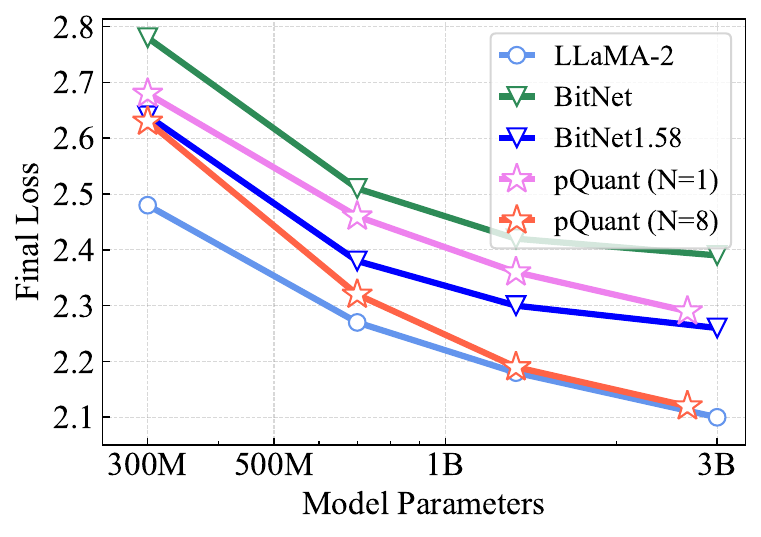}}
  \caption{Final training loss with varying numbers of parameters. pQuant with N=8 demonstrates excellent scalability, whereas 1-bit BitNet do not.}
  \label{fig:scaling_curve}
  \end{center}
\end{figure}

We scale pQuant by increasing the number of high-precision expert branches $N$ from 1 to 8, while maintaining the same number of active parameters per forward pass as a standard FFN. As shown in \autoref{fig:scaling_curve}, at $N=8$, pQuant consistently outperforms BitNet1.58 in training loss across all model sizes. More importantly, the performance gap between pQuant and the FP16 LLaMA-2 baseline narrows as model size increases: at 1.3B parameters, pQuant nearly matches LLaMA-2’s loss, whereas 1-bit BitNet and 2-bit BitNet1.58 exhibit a persistent gap. This contrast highlights a key finding: scaling via parameters is inherently inefficient under extreme quantization. In contrast, pQuant’s strategy of scaling high-precision branches enables it to follow a scaling law much closer to that of full-precision models, further evidencing its effective scalability. Downstream accuracy and perplexity results are reported in \autoref{tab:scaling_results_full}. The overhead introduced by adding more 8-bit branches remains minimal; a detailed analysis is provided in \autoref{appendix:Analysis of Scaling pQuant}.

\begin{figure*}[ht]
  \centering
  \subcaptionbox{Sensitivity analysis of pQuant}{\includegraphics[width = 0.6\linewidth]{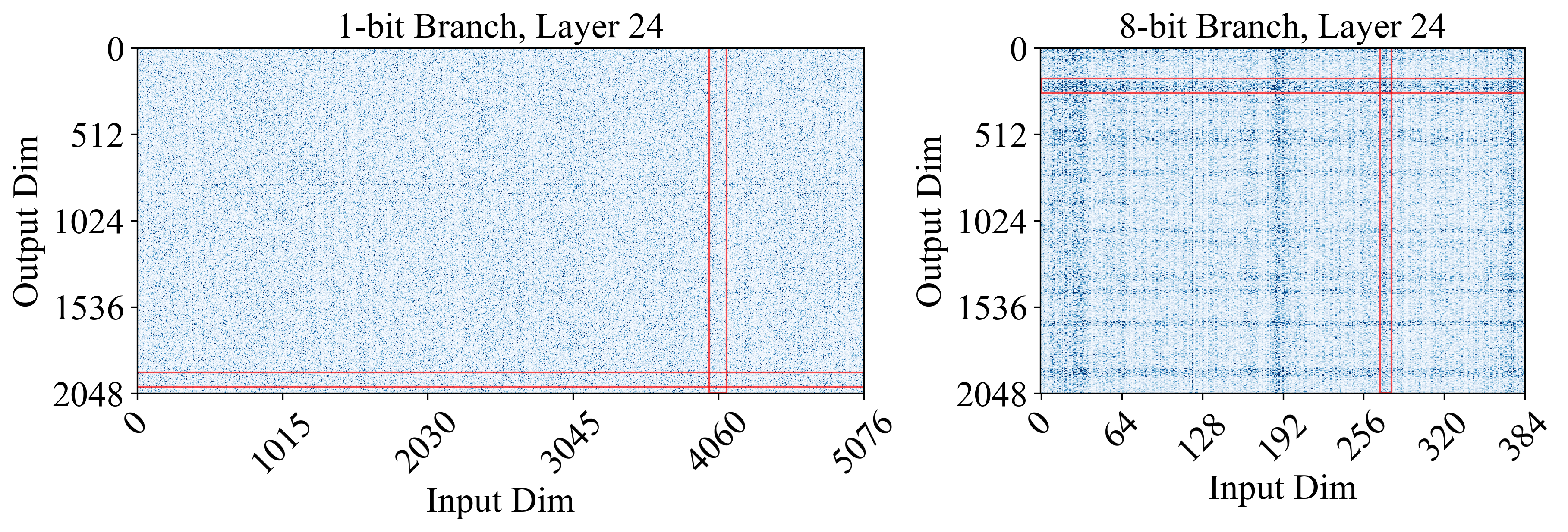}}
  \subcaptionbox{Ablation on pQuant}{\includegraphics[width = 0.38\linewidth]{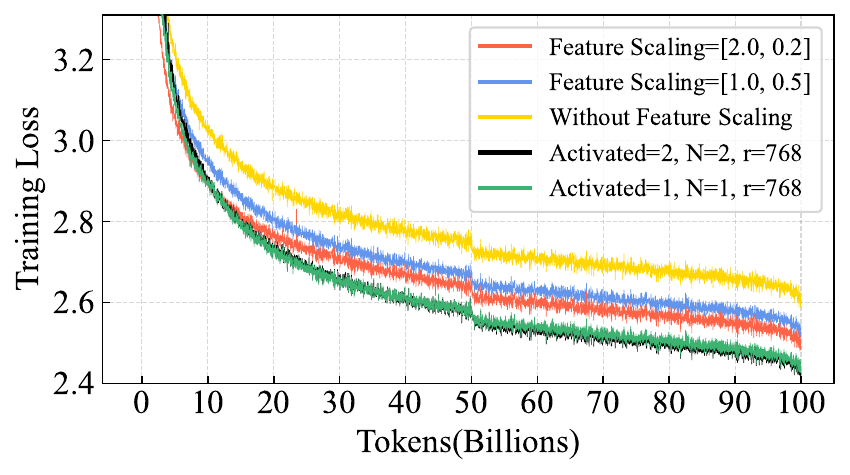}}
  \caption{(a) Sensitivity analysis of the 1-bit and 8-bit branches in the down-projection layer of the final FFN block in the 700M pQuant model. (b) Ablation study of pQuant on 700M-parameter model, assessing the impact of feature scaling and the number of active 8-bit branches on final loss. The mid-training loss drop stems from the two-phase learning rate schedule (see Appendix\ref{appendix:Two-Phase Training Schedule}).}
  \label{fig:Sensitivity Distribution and feature scaling} 
\end{figure*}

\subsection{Sensitivity Distribution of pQuant}
To validate that pQuant effectively mitigates parameter democratization, we analyze the weight sensitivity distribution in the final FFN layer of pQuant-1.3B after training, as shown in \autoref{fig:Sensitivity Distribution and feature scaling}(a). Compared to the uniform sensitivity distribution observed in BitNet-3B, pQuant exhibits a markedly differentiated sensitivity profile, with distinct regions of high and low sensitivity. This contrast confirms that pQuant’s decoupled architecture and feature scaling successfully guide the model to allocate sensitivity-critical parameters effectively, thereby enhancing expressivity and performance.

\subsection{Memory Efficiency of pQuant}
\label{Memory Efficiency}
\begin{figure}[ht]
  \begin{center}
  \centerline{\includegraphics[width=\columnwidth]{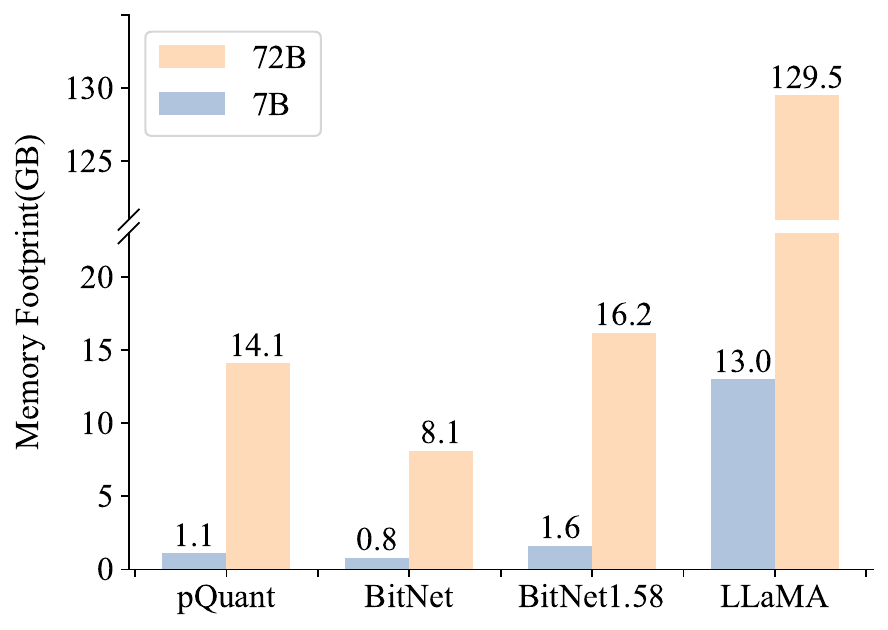}}
  \caption{The memory footprint. Memory footprint of model weights transferred during a single forward pass in inference. pQuant is smaller than BitNet1.58 and significantly lower than LLaMA-2.}
  \label{The memory footprint}
  \end{center}
\end{figure}
The process of transferring model parameters from DRAM to the memory of an on-chip accelerator (e.g., SRAM) can be expensive during inference ~\cite{yuan2024llm}. Compared to half-precision models, 1-bit models also suggests the opportunity to improve the serving latency. Thus, we outline the required memory for various model sizes in \autoref{The memory footprint}. pQuant maintains a consistent memory footprint during decoding regardless of the value of $N$, as it activates only a single 8-bit branch at any given time. This design avoids excessive memory accesses and better leverages the characteristics of modern hardware, where bandwidth is often the primary bottleneck rather than memory capacity. Compared to LlaMA-2, pQuant reduces memory usage by 92\%, and requires 31\% less memory than BitNet1.58. In pQuant, numerous individual scalars are introduced: $alpha$, $beta$, $gamma$, $mu$, and $lambda$. These scalars can be merged during inference, ensuring that memory efficiency is not impacted.

\subsection{Ablation Study}
\label{Ablation Study}
\paragraph{Feature Scaling.}
As show in \autoref{fig:Sensitivity Distribution and feature scaling} (b), we initialize the learnable factory as $\alpha = 1.0$ and $\beta = 0.5$. Post-training analysis reveals distinct converged values (mean $\alpha \approx 2.0$, $\beta \approx 0.2$); re-initializing to these values improves performance. Crucially, models trained under different scaling configurations do \emph{not} converge to similar final losses, indicating that feature scaling exerts a persistent structural influence, rather than merely shaping early optimization dynamics. Ablating feature scaling leads to a significant increase in training loss, confirming its necessity for effective allocation of sensitivity-critical parameters. Further analysis is provided in \autoref{appendix:feature scaling}.

\paragraph{High-precision Branch.}
For hardware efficiency, we restrict the 8-bit branch dimension $r$ to integer multiples of 128. First, under a single-branch setting, increasing $r$ from 256 to 768 (at 700M scale) improves performance, but the gains do not compensate for the added memory and computation overhead. Second, we evaluate a multi-branch strategy that activates two branches of size $r=768$ per token, which yields negligible improvement.

In our experiments, pQuant with $N=4$ achieves performance better than BitNet1.58. To rigorously evaluate the architectural efficiency of pQuant, we conduct a controlled comparison under matched parameter budgets. Specifically, we reduce the hidden state dimension of pQuant to match the total of 1.3B parameters in BitNet1.58. We configure pQuant with $N=8$ and 926M activated parameters, using only one active branch. Under these settings, pQuant runs 1.6$\times$ faster than BitNet1.58 during inference. The results are shown in \autoref{Comparison under Same Total Parameters}. When evaluated on the same test set as our main experiments, both models achieve comparable performance. This parity in results, achieved despite pQuant's reduced hidden dimension, provides compelling evidence for pQuant's superior parameter efficiency and architectural advantages in low-bit quantization scenarios.

Additionally, we evaluated the practical performance using FP16 for 8-bit parameters and found that the loss curve closely matches that of the setting with one activated 8-bit branch and $r=576$, indicating that 8-bit precision, combined with a scaling factor, is already sufficient to represent the sensitive parameters in our architecture.

\begin{figure}[t]
\begin{center}
\centerline{\includegraphics[width=\columnwidth]{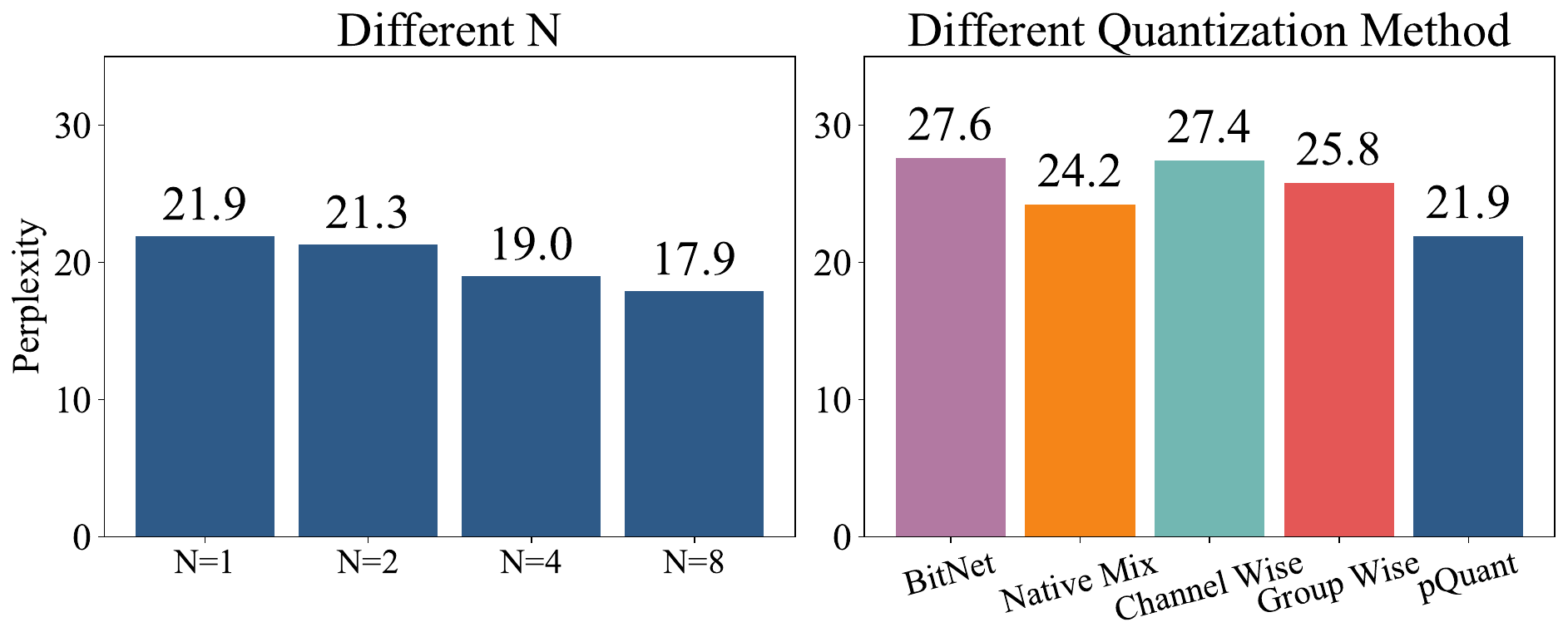}}
\caption{Perplexity of pQuant under different configurations. Lower perplexity indicates better performance.
Left: Perplexity decreases as the number of 8-bit branches N in pQuant increases.
Right: Comparison with alternative quantization methods. Native Mix retains 8\% of parameters in high precision on top of 1-bit BitNet, rather than using our decoupled architecture. Channel-wise quantizes weights per output channel of the matrix multiplication. Group-wise quantizes weights in groups of 64 elements.}
\label{N}
\end{center}
\end{figure}

We show the number of 8-bit branches $N$ ranging from 1 to 8, at 700M in the left of \autoref{N}. As $N$ increases, the performance of pQuant improves steadily. Specifically, in our tests, pQuant outperformed 2-bit model when $N=4$. Moreover, as shown in the right panel of \autoref{N}, we explored directly keeping 8\% of parameters in high precision from the start of BitNet training. This approach yields significantly worse performance than pQuant, despite using more high-precision parameters (8\% vs. pQuant’s 5\%). We also evaluated channel-wise and group-wise quantization. Channel-wise quantization provides negligible gains. Group-wise quantization (with block size 64) achieves better accuracy but requires storing one 16-bit scaling factor per 64 1-bit weights, which incurs substantial metadata overhead and is unfriendly to hardware. These results highlight the effectiveness of pQuant.

Additional ablation studies are provided in \autoref{appendix:Ablation Study}.

\begin{table}[t]
  \begin{center}
  \begin{adjustbox}{width=\linewidth}
  \begin{tabular}{cccccc}
  \toprule
  \textbf{Model} & $\textbf{Total}$ & $\textbf{Activated}$ & $\textbf{PPL}\downarrow$ & $\textbf{Memory}$\\
  \midrule
  pQuant($N=4$) & 1.5B & 1.3B & 15.5 & 0.91G \\
  BitNet1.58 & 1.3B & 1.3B & 16.9 & 0.72G \\
  pQuant($N=8$) & 1.3B & 926M & 16.8 & 0.98G \\
  LlaMA-2 & 1.3B & 1.3B & 14.4 & 2.64G \\
  \bottomrule
  \end{tabular}
  \end{adjustbox}
  \end{center}
  \caption{Performance comparison under Matched Parameter. Memory footprint include the storage of Embeddings and LayerNorm.}
  \label{Comparison under Same Total Parameters}
  \end{table}
  
  % 3*96*896*24*8 

\subsection{Limitations}
\label{Limitations}

Although QAT-Scratch methods, including pQuant, gain significant advantages in model performance, they generally incur higher training costs compared to traditional QAT and PTQ. The training duration across different models is presented in \autoref{appendix:Training Time}.

While pQuant can achieve better inference speed and model performance than 2-bits BitNet1.58 when $N\geq4$, this configuration introduces higher physical memory requirements, as shown in \autoref{Overhead of Model Parameter}. The speed improvement stems primarily from reduced memory access volume. Our design aligns with current trends in edge-device hardware development, where memory capacity is more cost-effective than bandwidth. This memory overhead represents a meaningful trade-off that merits further investigation.  

Due to the training cost, our experiments are limited to model size. Future work could explore scaling pQuant to larger models (e.g., 70B) to further validate its effectiveness.

\section{Conclusion}
\label{sec:conclusion}
We propose pQuant, a quantization-aware training-from-scratch method for extremely low-bit LLMs. Motivated by the observation that existing 1-bit models suffer from \emph{parameter democratization}, i.e., the homogenization of weight sensitivity that limits expressivity and scaling, we design a decoupled linear layer that preserves sensitive weights in a small high-precision branch. Coupled with feature scaling to prioritize gradient flow toward informative components, this approach enables effective allocation of model capacity under extreme quantization.  Experiments show that pQuant not only surpasses prior extremely low-bit methods at matched scales but also scales effectively.

\bibliography{custom}

\clearpage
\appendix

\section{Inference}
\label{appendix:Inference}
Although pQuant introduces additional operations such as quantization, de-quantizaiton, rescale and layernorm during training that slightly increase training overhead, these costs are entirely eliminated during inference, leading to highly efficient deployment. First, the parameters in 1-bit branch are offline quantized and stored in 1-bit precision during inference, matching the effect of PTQ methods. They are packed into UINT8 format with 8 parameters per byte, resulting in a storage footprint that is 1/16 of that required by FP16 models. This also reduces memory bandwidth when GPU cores load weight matrices from global memory. Second, the scaling factors used during training can be fused into preceding or succeeding layers during inference. For example, per-tensor scaling factors can be integrated into the input normalization of the subsequent module. Similarly, the RMSNorm operation can be merged with activation quantization, as both are element-wise transformations. These optimizations eliminate the need for half-precision in GEMM (General Matrix Multiplications) operations, allowing us to employ mixed-precision operators that support W1A8 GEMM computations within linear ~\cite{ladder}. Third, in the first GEMM of the FFN layer during inference, the same input must be multiplied with both the 8-bit and 1-bit branches of the up projection. To improve efficiency, this computation can be distributed across multiple thread groups, enabling parallel execution without redundant data reads.

In pQuant, the acceleration method is built atop T-MAC~\cite{wei2024tmaccpurenaissancetable}, a CPU-based inference library that natively supports mixed-precision matrix multiplication without dequantization. 1-bit quantization is particularly valuable in edge-device deployment, where the batch size is typically one and the most time-consuming operation becomes GEMV (General Matrix-Vector Multiplication) rather than GEMM. A promising approach to implement bitwise computation is through table lookups ~\cite{park2024lut}. Since each 1-bit weight can only take two values, the number of possible bit patterns in a small group is limited. For instance, if a 1-bit matrix is partitioned into groups of four elements, there are only $2^4$ possible combinations per group (e.g., [1,1,1,-1], [1,1,-1,-1]). Given an activation vector, the results of its multiplication with all possible bit patterns can be precomputed and stored in a lookup table. The mixed-precision GEMM between activation and 1-bit weights is then transformed into a series of table lookups indexed by the bit patterns in the weight, followed by summation of the retrieved results. This reduces GEMM to table lookup and addition operations, eliminating multiplications entirely.

\begin{figure}[h]
    \begin{center}
    \centerline{\includegraphics[width=\columnwidth]{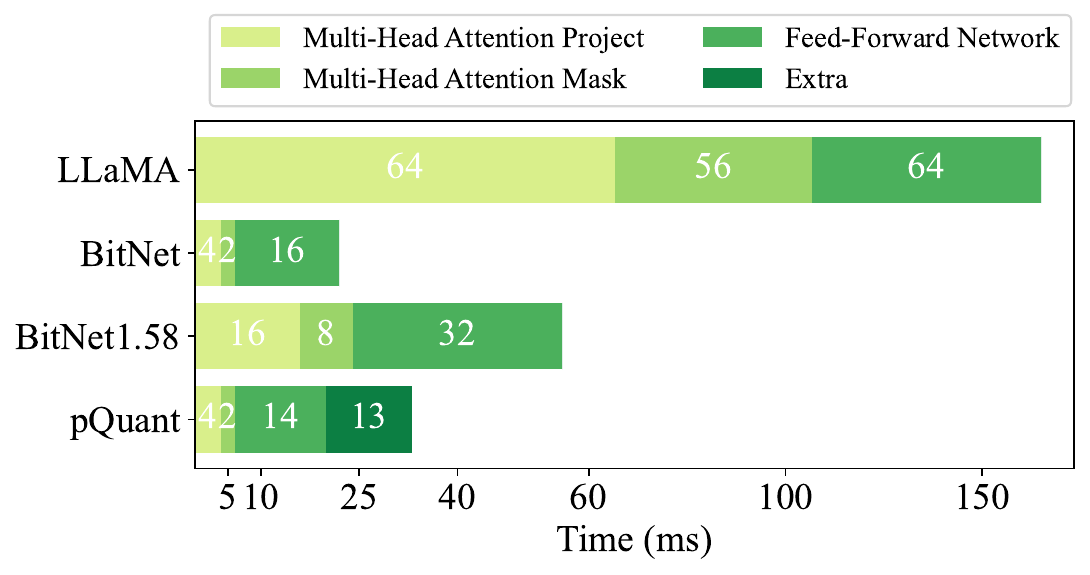}}
    \caption{Computation time across components in a Transformer block, measured on an Apple M2 with a 7B model and input sequence length of 256. pQuant is 38\% faster than BitNet1.58 and 82\% faster than FP16 LLaMA-2.}
    \label{The computation time cost}
    \end{center}
\end{figure}

\paragraph{Computation Efficiency} 
The use of 1-bit weights eliminates a significant amount of matrix multiplication operations, thereby substantially reducing computational complexity~\cite{zhu2024scalable}. To further evaluate its practical efficiency, we conducted comprehensive benchmarking of overhead during inference. \autoref{The computation time cost} presents the time costs for linear operation across different models. Compared to BitNet1.58 and LlaMA-2, pQuant demonstrates an 82\% improvement in computational performance over LlaMA-2 and is notably more efficient than BitNet1.58, reducing computation time by 38\%.

\section{Training Mechanism}
\label{appendix:training}

We introduce RMSNorm to compress the dynamic range of activations. This helps maintain basic performance under absmean-based quantization and has been shown to accelerate convergence in 1-bit quantized models ~\cite{ma2024era, tiionebitllms}. This provides a smoother optimization landscape for gradient descent and improves convergence ~\cite{liu2023oscillation}. The data is preprocessed using the BPE tokenizer with a vocabulary size of 32K. We select 500 steps for warm-up.

\subsection{Straight-Through Estimator} 
Since the function sign(·) is non-differentiable at zero, it causes the gradient chain to break during back propagation, preventing optimization of the model parameters. Therefore, we use the Straight-Through Estimator (STE) method during back propagation. In traditional back propagation, gradients flow through differentiable functions allowing for the optimization of weights via gradient descent. However, when operations are not continuous or involve discrete decisions, calculating these gradients becomes problematic. The STE provides a way around this by passing the gradient straight through the operation during the backward pass while using the quantized values in the forward pass ~\cite{li2024continuous}. In essence, STE approximates the gradient of a non-differentiable function as 1, allowing the network to learn despite the presence of non-differentiable operations.

\begin{figure}[ht]
  \centering
  \centerline{\includegraphics[width=0.95\linewidth]{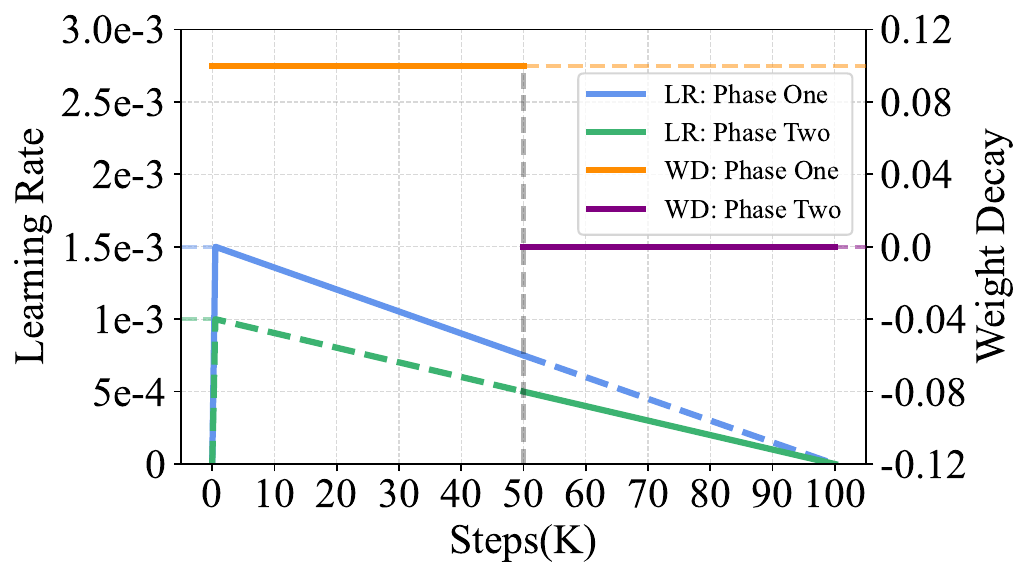}}
  \caption{Two-Stage Scheduler. During mid-training, both the learning rate (LR) and weight decay (WD) are decayed to accelerate convergence of 1-bit parameters and prevent sign flips caused by excessive weight oscillation around quantization thresholds.}
  \label{Learning Rate and Weight Decay}
\end{figure}

\subsection{Two-Phase Training Schedule}
\label{appendix:Two-Phase Training Schedule}
Besides, pQuant employ a two-phase training schedule ~\cite{bitnetfaq}. In the first phase, the learning rate begins at a high value and decreases linearly until the midpoint of training. The second phase starts with a lower learning rate, which also decays linearly. Besides, weight decay is set to 0.1 during the first phase and disable in the second phase. These strategies have been shown to improve convergence in low bit training, as demonstrated by BitNet1.58. This approach results in a significant reduction in the loss curve shortly after the midpoint. The learning rates and weight decay for the two phases are shown in the \autoref{Learning Rate and Weight Decay}.

There are two primary motivations for this design. First, in contrast to FP16 models where learning rates are typically well calibrated, 1-bit optimization faces a key challenge: small updates to the latent weights often fail to produce meaningful changes in the quantized 1-bit weights. This results in biased gradient estimation and suboptimal update behavior, particularly during the early training stages when rapid convergence is critical. Second, in half-precision training, weight decay serves as an effective regularizer by constraining large weight magnitudes, thereby preventing overfitting and improving training stability. However, in 1-bit training, weight decay is applied to the latent weights rather than the actual 1-bit weights, which alters its regularization effect and may lead to unintended consequences. 

\section{Experiment Details}
\label{Experiment Details}
These common sense reasoning benchmarks we chose in our experiments have been widely adopted in prior work to assess quantization effectiveness, in both post-training quantization (PTQ) and quantization-aware training (QAT) methods ~\cite{xiao2023smoothquant, lin2024awq, chee2024quip,shang2023pb, ashkboos2024quarot}. We also use perplexity to quantitatively measure the model’s generation power on WikiText-2~\cite{merity2016pointer}. We followed the pipeline from lm-evaluation-harness4~\cite{eval-harness} to perform the evaluation. Furthermore, we directly mixed and shuffled the training datasets for training. For performance evaluation, each experiment was repeated 10 times and averaged. Due to high training costs, each model was trained only twice to mitigate errors—though in practice, loss curves under identical configurations were nearly identical. 

The configurations for the BitNet, BitNet1.58, and LlaMA-2 models are shown in \autoref{Baseline configurations}. The sequence length was uniformly set to 2048. To optimize training efficiency and performance, we employed mixed-precision training, and the Adam optimizer is applied with $\beta_1 = 0.9$ and $\beta_2 = 0.95$.

Our models were trained using 16 NVIDIA A100-80G GPUs and 1 TB of CPU memory. We utilized the DeepSpeed framework~\cite{rasley2020deepspeed} for training.

\begin{table}[ht]
\begin{center}
\begin{adjustbox}{width=0.85\linewidth}
\begin{tabular}{ccccc}
\toprule
\textbf{Parameters} & $\textbf{D}_{\textbf{Model}}$ & $\textbf{D}_{\textbf{FF}}$ & \textbf{Heads} & \textbf{Layers} \\
\midrule
300M & 1024 & 2400 & 16 & 24 \\
700M & 1536 & 4096 & 24 & 24 \\
1.3B & 2048 & 5460 & 32 & 24 \\
\bottomrule
\end{tabular}
\end{adjustbox}
\end{center}
\caption{Configurations for BitNet, BitNet1.58 and LlaMA-2. Head denotes the number of attention heads in multi-head attention; Layers denotes the number of Transformer blocks. }
\label{Baseline configurations}
\end{table}
  
\section{Analysis of Scaling pQuant}
\label{appendix:Analysis of Scaling pQuant}
The main results in \autoref{tab:eval_results} suggests that merely increasing the parameter count is insufficient for extracting additional information in the extremely low-bit scenario. The expressive capacity of 1-bit tensor is highly constrained under the absmean quantization scheme, limiting its scalability.

The \autoref{tab:scaling_results_full} below presents the complete performance test results after scaling.
\begin{table*}[ht]
\begin{center}
\begin{adjustbox}{width=\textwidth}
\begin{tabular}{ccccccccccc}
\toprule
\textbf{Models} & \textbf{Total/Activated} & \textbf{ARC-E$\uparrow$} & \textbf{ARC-C$\uparrow$} & \textbf{HS$\uparrow$} & \textbf{BQ$\uparrow$} & \textbf{OQ$\uparrow$} & \textbf{PQ$\uparrow$} & \textbf{WGe$\uparrow$} & \textbf{Avg.$\uparrow$} & \textbf{Perplexity}$\downarrow$ \\
\midrule
LlaMA-2 (16-bit) & 300M & \textbf{44.8} & 19.3 & \textbf{34.8} & \textbf{60.7} & \textbf{18.6} & \textbf{66.8} & 51.3 & \textbf{42.3} & \textbf{22.9} \\
BitNet1.58 (2-bit) & 300M & 43.8 & 19.7 & 34.6 & 57.5 & 16.0 & 64.9 & 52.2 & 41.2 & 29.7 \\
\textbf{pQuant(N=8)} & 366M/300M & 43.5 & \textbf{20.1} & 34.5 & 60.2 & 17.2 & 64.6 & \textbf{51.9} & 41.7 & 27.1 \\
\midrule
LlaMA-2 (16-bit) & 700M & \textbf{51.4} & \textbf{21.4} & \textbf{37.3} & \textbf{58.9} & \textbf{19.0} & 68.5 & 52.4 & \textbf{44.1} & \textbf{16.5} \\
BitNet1.58 (2-bit) & 700M & 48.5 & 20.0 & 35.9 & 57.7 & 17.8 & 68.3 & 51.1 & 42.8 & 21.1 \\
\textbf{pQuant(N=8)} & 898M/700M & 48.5 & 20.7 & 36.7 & 58.6 & 18.6 & \textbf{68.5} & \textbf{52.6} & 43.5 & 17.9 \\
\midrule
LlaMA-2 (16-bit) & 1.3B & \textbf{53.9} & \textbf{22.1} & 38.7 & 55.7 & \textbf{21.8} & \textbf{71.0} & 54.3 & 45.4 & \textbf{14.4} \\
BitNet1.58 (2-bit) & 1.3B & 50.4 & 21.2 & 36.9 & \textbf{61.2} & 19.4 & 68.7 & 53.0 & 44.4 & 16.9 \\
\textbf{pQuant(N=8)} & 1.7B/1.3B & 52.9 & 21.7 & \textbf{38.9} & 60.0 & 21.6 & 70.3 & \textbf{55.5} & \textbf{45.8} & 14.3 \\
\bottomrule
\end{tabular}
\end{adjustbox}
\caption{Average score on downstream tasks. Results show that pQuant achieves efficient scaling by increasing the capacity of the high-precision branch. When the number of 8-bit branch reaches 8, pQuant surpasses 2-bit baselines and matches the accuracy of FP16 models. Total/Activated means total parameters/activated parameters per forward pass.}
\label{tab:scaling_results_full}
\end{center}
\end{table*}

\subsection{Overhead of Model Parameter}
\label{Overhead of Model Parameter}
Although the number of activated parameters during inference in pQuant is consistent with that of other models, the overall parameter count increases with varying N values. Specifically, the total parameter counts for models of different sizes, when N is 1, 2, 4 or 8 are shown in the \autoref{Total Parameter}. When N=1, pQuant has the same number of parameters as normal transformers model.

\begin{table}[ht]
\begin{center}
\begin{adjustbox}{width=0.8\linewidth}
\begin{tabular}{cccccc}
\toprule
\textbf{Base Size} & $\textbf{N=1}$ & $\textbf{N=2}$ & $\textbf{N=4}$ & $\textbf{N=8}$ \\
\midrule
300M & 300M & 309M & 328M & 366M \\
700M & 700M & 728M & 785M & 898M \\ 
1.3B & 1.3B & 1.4B & 1.5B & 1.7B \\  
\bottomrule
\end{tabular}
\end{adjustbox}
\end{center}
\caption{Total parameters of pQuant. N denotes the number of 8-bit branches used.}
\label{Total Parameter}
\end{table}

\section{Ablation Study}
\label{appendix:Ablation Study}

\paragraph{Batch Size.} We evaluated various batch sizes and found 1 million tokens led to significant performance improvements compared to 4 million tokens. This indicates that low-bit models benefit more from frequent parameter updates to achieve optimal performance, rather than relying on larger batch sizes for stability.

\paragraph{Learning rate.} As illustrated in \autoref{fig:Sensitivity Distribution and feature scaling}(b), we observed a significant reduction in loss occurring when the learning rate was decayed, rather than at the end of the training phase. This approach made the performance more predictable during the training process. However, half-precision models did not benefit from a similar learning rate decay strategy, likely because their loss curves do not exhibit an S-shaped pattern.

\section{Feature Scaling Analysis}
\label{appendix:feature scaling}
As illustrated in \autoref{final feature_scaling}, the feature scaling values of 1.3B pQuant at the end of training exhibit two clear patterns: (1) both the initial and final layers exhibit relatively large scaling magnitudes, while the intermediate layers show smaller values. (2) the scaling factors for the 8-bit branches are significantly higher than those for the 1-bit branches, which aligns with our design principle of allocating higher precision to more influential weights.

\begin{table*}[ht]
  \begin{center}
  \begin{adjustbox}{width=\textwidth}
  \begin{tabular}{ccccccccccccccccc}
  \toprule
  \textbf{Layers} & $\textbf{1}$ & $\textbf{2}$ & $\textbf{3}$ & $\textbf{4}$ & $\textbf{6}$ & $\textbf{8}$ & $\textbf{10}$ & $\textbf{12}$ & $\textbf{14}$ & $\textbf{16}$ & $\textbf{18}$ & $\textbf{20}$ & $\textbf{21}$ & $\textbf{22}$ & $\textbf{23}$ & $\textbf{24}$ \\
  \midrule
  8-bit   & 1.82 & 1.73 & 1.57 & 1.59 & 1.39 & 1.33 & 1.25 & 1.22 & 1.34 & 1.49 & 1.37 & 1.27 & 1.38 & 1.45 & 1.56 & 1.61 \\
  1-bit   & 0.78 & 0.62 & 0.48 & 0.42 & 0.38 & 0.34 & 0.31 & 0.26 & 0.29 & 0.36 & 0.43 & 0.45 & 0.41 & 0.47 & 0.52 & 0.50 \\
  \bottomrule
  \end{tabular}
  \end{adjustbox}
  \end{center}
  \caption{Feature scaling after training. In the trained model, feature scaling reveals the model’s preference for the 8-bit branch, highlighting its ability to preserve outliers.}
  \label{final feature_scaling}
  \end{table*}

\section{Stability of pQuant}
\label{appendix:Stability of pQuant}

\begin{figure}[ht]
  \begin{center}
  \centerline{\includegraphics[width=0.9\columnwidth]{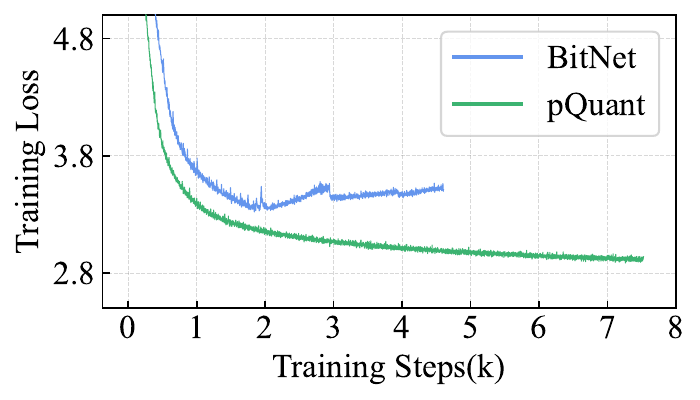}}
  \caption{Training loss curve of two model. pQuant exhibits greater training stability. In contrast, BitNet frequently suffers from gradient explosion during training, often requiring checkpoint reloading and restarts.}
  \label{stable}
  \end{center}
\end{figure}

Additionally, we observed that the 1-bit model experienced instability and crashes at a batch size of 4 million tokens, which is the batch size used for LlaMA-2. Even under the optimal batch size of 1M suggested in the BitNet research, training often becomes unstable. As shown in the \autoref{stable}, this necessitates frequent rollbacks to previous checkpoints to resume training. In contrast, pQuant did not show similar issues under the same conditions, indicating its superior stability.

\section{Training Time}
\label{appendix:Training Time}
The gradient update process in QAT models differs from that in traditional pre-training, often involving a simpler or more constrained optimization landscape. pQuant follows this characteristic. This is primarily due to the additional computational steps involved in quantization, dequantization, and auxiliary matrix operations, which are currently implemented using high-precision arithmetic. The consequence of latent weight keeping during training is that we are keeping both a FP16 and 1-bit copy of weights during training. This makes training less memory efficient than in a standard Transformer. Since large models are bottlenecked on memory bandwidth, this has a significant impact on training performance.

The estimated training duration across different models is presented in \autoref{Total days for training}.
\begin{table}[ht]
\begin{center}
\begin{adjustbox}{width=0.75\linewidth}
\begin{tabular}{cccccc}
\toprule
\textbf{Parameters} & $\textbf{N=1}$ & $\textbf{N=2}$ & $\textbf{N=4}$ & $\textbf{N=8}$ \\
\midrule
300M & 1.9 & 2.0 & 2.1 & 2.3 \\
700M & 4.9 & 5.1 & 5.4 & 6.0 \\
1.3B & 8.5 & 8.8 & 9.6 & 11.1 \\
\bottomrule
\end{tabular}
\end{adjustbox} 
\caption{Total days for training.}
\label{Total days for training}
\end{center}
\end{table}

\section{Related Work}

\subsection{Post-Training Quantization}
\label{appendix:Post-Training Quantization}
Current quantization approaches are broadly categorized into Quantization-Aware Training and Post-Training Quantization. QAT incorporates quantization and dequantization during the training phase, enabling the model to adapt to the constraints imposed by reduced precision. In contrast, PTQ applies quantization without requiring further training. 

A number of recent studies ~\cite{luo2020positional, bondarenko2021understanding, wei2023outlier, xiao2023smoothquant} have identified the existence of significant outliers in large language models. Outlier Suppression (OS) ~\cite{wei2022outlier} restructures the LayerNorm function by shifting its scale parameter $\gamma$ to reduce outlier effects. OS+ ~\cite{wei2023outlier} introduces a channel-wise shifting transformation that balances activation magnitudes across channels to handle asymmetric distributions. SmoothQuant ~\cite{xiao2023smoothquant} applies a per-channel scaling transformation that redistributes quantization difficulty from activations to weights. AWQ ~\cite{yao2021hawq} observes that the importance of weight channels is largely determined by activation scales. The idea of rotation transformation was first explored in QuIP ~\cite{chee2024quip}, later extended by QuaRot ~\cite{ashkboos2024quarot}, which integrates quantization for both weights and activations, including the KV cache. GPTQ ~\cite{frantar2022gptq} adopts a more accurate quantization framework and achieves strong results in 4-bit settings.

SpinQuant ~\cite{liuspinquant} finds that different rotation matrices can lead to varying performance in quantized models and proposes a learning-based approach to optimize rotation transformations. ZeroQuant ~\cite{yao2022zeroquant} and LLM.int8() ~\cite{dettmers2022gpt3} improve quantization accuracy by introducing custom grouping strategies for quantization blocks. SpQR ~\cite{dettmers2023spqr} further partitions features to better manage precision constraints. OWQ ~\cite{lee2024owq} preserves information capacity in sensitive weight channels through carefully designed scaling transformations.

\subsection{Quantization-Aware Training}
\label{appendix:Quantization-Aware Training}
\begin{figure}[ht]
  \centering
  \centerline{\includegraphics[width=0.95\linewidth]{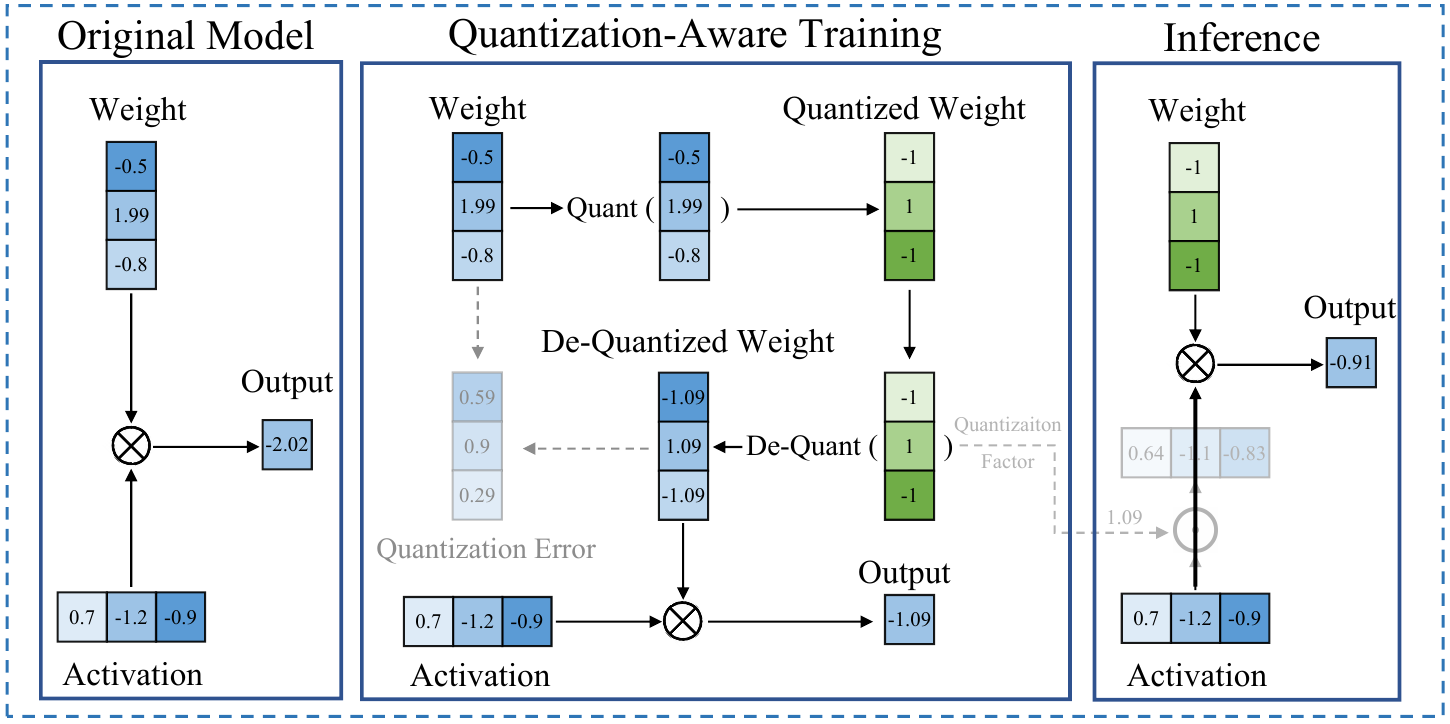}}
  \caption{Illustration of the Quantization-Aware Training process. Before quantization, matrix multiplications are performed in 16-bit precision. During QAT, simulated quantization are inserted to enable the model to learn and compensate for quantization-induced errors. In the final inference phase, quantization factors are fused into the weights, only use the quantized weights. }
  \label{Quantization_error}
\end{figure}

In QAT, the training process typically involves three key components: quantization simulation, gradient approximation, and parameter optimization. During forward propagation, weights are passed through a quantize-dequantize pipeline to mimic the effect of low bit width computation in real deployment. Specifically, they are first quantized into lower precision representations and then dequantized back to higher precision values for further computation. Since quantization operations such as rounding are non-differentiable ~\cite{yin2019understanding}, gradient approximation techniques are employed during back propagation. With this setup, the model learns to optimize its parameters under simulated quantization conditions, thereby improving robustness to quantization noise ~\cite{shen2021once}. As illustrated in \autoref{Quantization_error}, given a 16-bit weight matrix, a quantization scale is typically derived from its dynamic range (e.g., maximum absolute value). The matrix is then scaled, rounded to 1-bit integers, and dequantized back to floating point. The difference between the dequantized and original matrices constitutes the quantization error. 

The effectiveness of QAT largely depends on the design of the pseudo-quantization operation, which varies in how well it reduces model sensitivity to quantization. A well-designed pseudo-quantization mechanism allows the model to adjust its weights during training in response to quantization effects, making it more robust or "numb" to the distortions introduced during actual quantization.

\end{document}